# Enhancing Cooperative Coevolution for Large Scale Optimization by Adaptively Constructing Surrogate Models

Bei Pang
Autocontrol Institute, Xi'an Jiaotong University
Xi'an, Shaanxi, 710049, P. R. China
beibei@stu.xjtu.edu.cn

Zhigang Ren
Autocontrol Institute, Xi'an Jiaotong University
Xi'an, Shaanxi, 710049, P. R. China
renzg@stu.xjtu.edu.cn

Yongsheng Liang
Autocontrol Institute, Xi'an Jiaotong University
Xi'an, Shaanxi, 710049, P. R. China
liangyongsheng@stu.xjtu.edu.cn

An Chen
Autocontrol Institute, Xi'an Jiaotong University
Xi'an, Shaanxi, 710049, P. R. China
chenan123@stu.xjtu.edu.cn

## ABSTRACT

It has been shown that cooperative coevolution (CC) can effectively deal with large scale optimization problems (LSOPs) through a divide-and-conquer strategy. However, its performance is severely restricted by the current context-vector-based sub-solution evaluation method since this method needs to access the original high dimensional simulation model when evaluating each sub-solution and thus requires many computation resources. To alleviate this issue, this study proposes an adaptive surrogate model assisted CC framework. This framework adaptively constructs surrogate models for different sub-problems by fully considering their characteristics. For the single dimensional sub-problems obtained through decomposition, accurate enough surrogate models can be obtained and used to find out the optimal solutions of the corresponding sub-problems directly. As for the nonseparable sub-problems, the surrogate models are employed to evaluate the corresponding sub-solutions, and the original simulation model is only adopted to reevaluate some good sub-solutions selected by surrogate models. By these means, the computation cost could be greatly reduced without significantly sacrificing evaluation quality. Empirical studies on IEEE CEC 2010 benchmark functions show that the concrete algorithm based on this framework is able to find much better solutions than the conventional CC algorithms and a non-CC algorithm even with much fewer computation resources.

## 1 INTRODUCTION

Large-scale optimization problems (LSOPs) are becoming more and more popular in scientific research and engineering applications [1]. Due to the black-box characteristics of this kind of problems, the gradient-free evolutionary algorithms (EAs) are often employed to tackle them. However, the performance of conventional EAs often rapidly deteriorates as the problem dimension increases. This is the so-called 'curse of dimensionality' [2, 3].

Taking the idea of 'divide-and-conquer', cooperative coevolution (CC) [4] provides a natural way for solving LSOPs. It first decomposes the original LSOP into several smaller and simpler sub-problems, and then solves the LSOP by cooperatively optimizing all the sub-problems with a conventional EA. It is understandable that decomposition plays a fundamental role in CC. A right decomposition can reduce the optimization difficulty without changing the optimal solution. Therefore, in recent years, most research efforts on CC were put into designing various kinds of decomposition algorithms, and by now several efficient decomposition algorithms have been developed [5].

By contrast, another important algorithmic operation in CC, the sub-problem optimization, which also affects much on the efficiency of CC, is neglected. It is known that CC mainly focuses on black-box LSOPs which have no explicit objective functions and generally evaluates their solutions by simulation. This means that all the sub-problems obtained through decomposition do not own separate or explicit objective functions. To evaluate the sub-solutions, now all the CC algorithms adopt a context-vector-based method [3]. This method takes a complete solution of the original LSOP as context vector. For a sub-solution to be evaluated, the method first inserts it into the corresponding positions in the context vector, and then achieves evaluation by indirectly evaluating the modified context vector with the simulation model of the original LSOP. But generally, a very limited number of solution simulations are allowed for a practical LSOP since even a single simulation is relatively time-consuming, then the simulation times assigned to each sub-problem are further reduced. With so little computation resource, it is challenging for a conventional EA to produce high-quality solutions for sub-problems. As a result, the performance of the final solution to the original LSOP can hardly be guaranteed. Under this condition, it is very significant to develop efficient sub-problem optimization and sub-solution evaluation methods to decrease the requirement of CC on the computation resource.

As explained above, the sub-problems obtained through decomposition can be regarded as small or medium scale computationally expensive black-box optimization problems. To deal with this kind of problems more efficiently, surrogate model assisted EAs (SAEAs) were developed [6, 7]. Their key idea is to construct a calculable surrogate model for the computationally expensive objective function and employ the surrogate model to evaluate solutions. Only some promising solutions selected by the surrogate model need to access the real objective function. By this means, the number of real evaluations can be greatly reduced. Based on these characteristics of CC and surrogate model, it is natural to introduce surrogate model into CC.

In recent years, several types of surrogate models, including polynomial regression (PR), support vector regression (SVR), Gaussian process (GP) regression, and radial basis function (RBF), were proposed [8]. Among these models, PR is easy to train but generally shows low estimation accuracy when the dimensionality of the problem is high, SVR is able to relieve the curse of dimension but has difficulty in tackling large scale samples, GP can fit complex response surface well but asks long training time and shows dramatic performance deterioration as the problem dimension increases. As for RBF, it is easy to train and is relatively robust to the change of problem dimension [8]. Due to the differences in model characteristics, it is necessary to select different surrogate models for the problems with different characteristics.

By fully considering the characteristic of each sub-problem in CC, this study proposes an Adaptive Surrogate Model Assisted CC framework named ASMCC which can adaptively construct surrogate models for the sub-problems with different characteristics. It is clearly that some sub-problems are more easily to solve than others in CC. For example, if the decision variables involved in a sub-problem are independent of each other, the sub-problem can be easily solved by decomposing it into a number of 1-diemnsioanl sub-problems and solving the obtained 1-dimensional sub-problems in a very simple way. Based on this characteristic of CC, ASMCC solves the separable sub-problems and the nonseparable sub-problems of CC in two different ways. For the separable sub-problems, they are first further divided into a number of 1-dimensional sub-problems, and then the 1-dimensional sub-problems are solved by a two-layer surrogate search process. And for the nonseparable sub-problems, they are solved by a specific SAEA. In ASMCC, the two-layer surrogate search process only adopts the original simulation model to evaluate the training points which are used to construct the surrogate models, and there are only some selected promising sub-solutions in SAEA need to be reevaluated by the original simulation model, so that the algorithm can obtain better solution with much fewer computation resources.

To show the efficiency of ASMCC, we implement a concrete ASMCC algorithm. To achieve that, it is necessary to specify the type of surrogate models and the optimizer for each type of sub-problem. In ASMCC, the two layer surrogate search process is used to solve the 1-dimensional sub-problems, so that we select PR model as the surrogate model. And for the SAEA, it is used to solve the higher dimensional nonseparable sub-problems so that we select RBF model as the surrogate model, and adopt success-history based adaptive differential evolution (SHADE) [9] as the basic EA.

The rest of this paper is organized as follows. Section 2 describes the proposed ASMCC algorithm in detail, including the process of the ASMCC algorithm, the two-layer PR search process and the RBF assisted EA (RBF-SHADE). Section 3 reports the experimental studies. Finally, conclusions are drawn in section 4.

## 2 DESCRIPTION OF THE ASMCC ALGORITHM

ASMCC solves the LSOP by optimizing the separable and nonseparable sub-problems in different ways. In our ASMCC algorithm, the separable sub-problems are further divided into several 1-dimensional sub-problems and solved by a two-layer PR search process, then the nonseparable sub-problems are optimized by RBF-SHADE with the rest of function evaluations (FEs). The procedure of the ASMCC algorithm is presented in Algorithm 1.

| **Algorithm 1**: ASMCC |
|---|
| 1. Generate a decomposition $x \to \{x_1, x_2, \cdots, x_k\}$; |
| 2. Initialize the context vector $x^c$ with a randomly generated complete solution; |
| 3. **for** each 1-dimensional sub-problem $g$ **do** |
| 4.    $(x_g^*) = PR(g)$; |
| 5. Initialize the best overall solution $x^*$ based on the obtained best solutions of the 1-dimensional sub-problems; |
| 6. Initialize the parameters, population $P_g$ and database $D_g^n$ for all of the nonseparable sub-problems; |
| 7. **while** the termination condition is not met **do** |
| 8.    Determine the sub-problem $g$ to be optimized; |
| 9.    $(P_g, D_g^n, x^*) \leftarrow RBF-SHADE(P_g, D_g^n, x^*, g)$; |
| 10. Output $x^*$. |

In ASMCC, step 1 decomposes the original LSOP into several sub-problems. Within these sub-problems, the separable sub-problems have been further decomposed into a number of 1-dimensional sub-problems. Then, steps 2-4 optimize the 1-dimensional sub-problems with the two-layer PR search process, where a fixed context vector $x^c$ is used for simplicity (step 2).

After the separable sub-problems are solved, the nonseparable sub-problems are optimized by RBF-SHADE with the rest of computation resources (steps 5-9). In the process of solving the nonseparable sub-problems, the best overall solution $x^*$ is first initialized (step 5) and it is used as the context vector of RBF-SHADE. In $x^*$, the sub-vectors corresponding to the separable sub-problems are initialized with the optimal solutions found by the two-layer PR search process. And the sub-vectors corresponding to the nonseparable sub-problems are initialized randomly. Then, the parameters of RBF-SHADE, the population $P_g$ and the database $D_g^n$ which stores the training points of the RBF model of each nonseparable sub-problem are generated (step 6). Step 8 selects the nonseparable sub-problems with a round-robin method, and the sub-problem $g$ is just allowed to evolve one generation when it is selected. Steps 7-9 will repeat until the computation resources are exhausted.

In the following, we will describe the two-layer PR search process and RBF-SHADE in detail.

## 2.1 The Two-Layer PR Search Process

In this study, PR model is selected as the surrogate model of the 1-dimensional sub-problems. We directly use the function "*polyfit*" in MATLAB to construct our PR models, where $P = polyfit(X, Y, N)$ finds the coefficients $P$ of a polynomial $P(x)$ of degree $N$ that fits the data $Y$ best in a least-squares sense.

---

**Algorithm 2**: $(x_g^*) = PR(g)$

// the first layer
1. Initialize $D_g^s$ with $d^s$ uniformly generated sub-solutions $x_g$ within $[lb_g, ub_g]$ and evaluate them with $e(x_g)$;
2. Initialize the best solution of the $g$th sub-problem $x_g^*$;
3. **if** $|FDC_g| > \varepsilon$, $N=2$; **else** $N=5$; **end**
4. Build $P(x_g)$ with $D_g^s$;
5. $x_g^* = \text{fminbnd}(-P(x_g), lb_g, ub_g)$;
6. Define the new search region $[lb_g', ub_g']$;

// the second layer
7. Find out the solutions within $[lb_g', ub_g']$ from $D_g^s$ and store them into $O_g$;
8. Uniformly generate $d^s - |O_g|$ sub-solutions $x_g$ within $[lb_g', ub_g']$, store them into $Q_g$ and evaluate them with $e(x_g)$;
9. $D_g^s \leftarrow O_g \cup Q_g$;
10. Divide $[lb_g', ub_g']$ into $\lfloor d^s/6 \rfloor$ sub-regions;
11. Allocate each $x_g \in D_g^s$ into corresponding sub-region;
12. **for** sub-region $i=1:\lfloor d^s/6 \rfloor$ **do**
13.     Build $P_i(x_g)$ based on $D_g^i$, where $N=5$;
14.     $x_g^{i^*} = \text{fminbnd}(-P_i(x_g), lb_i, ub_i)$;
15. $i = \arg\max_{i=1:\lfloor d^s/6 \rfloor}(P_i(x_g^{i^*}))$;
16. **if** $f(x^c | x_g^{i^*}) < f(x^c | x_g^*)$ **then** $x_g^* \leftarrow x_g^{i^*}$;
17. Return $x_g^*$;

---

In the two-layer PR search process, the aim of the first layer is to find out a small region which covers the optimal solution of the sub-problem by constructing a global PR model, and the aim of the second layer is to find out the final optimal solution of the sub-problem within the small region by constructing several local PR models. It can be easily discovered that the first layer plays an important role in the two-layer search process because if it finds a false small region, the real optimal solution will not be found. The procedure of the two-layer PR search process used to solve the $g$th 1-dimensional sub-problem is described in Algorithm 2.

In the two-layer PR search process, the real fitness improvement value made by a new solution $x_g$ of a sub-problem to the context vector $x^c$ is adopted as the training point to construct PR model. The reason is that the indirect evaluation values $f(x^c | x_g)$ of sharply different sub-solutions $x_g$ to the same sub-problem may have small differences since each $f(x^c | x_g)$ adds up the fitness values of the solutions of all the sub-problems, which makes against constructing an accurate PR model. For a minimization problem, the fitness improvement of a sub-solution $x_g$ is defined as $e(x_g) = f(x^c) - f(x^c | x_g)$. For an additively separable problem, it eliminates the influence of other sub-problems and enlarges the relative differences among the evaluation values of different solutions to the same sub-problem. Obviously, a sub-solution making larger fitness improvement is considered better. In the following, we will describe the first and second layer of the two-layer PR search process in detail.

**The first layer:** In the first layer, database $D_g^s$ which records the training points of the PR model is first initialized with $d^s$ uniformly generated sub-solutions (step 1). Then, the fitness distance correlation (FDC) [10, 11] is used to determine the degree of the PR model (step 3). FDC is a kind of landscape analysis techniques for quantitatively measuring the difficulty of optimization problems. We can use FDC to acquire information about the degree of nonlinearity of the problem landscape for selecting a degree of PR model. Generally, low-order nonlinearity of fitness landscape will be shown if FDC coefficient is close to the value of 1 or −1, which indicates that the problem landscape is smooth, while high-order nonlinearity can be found if the FDC coefficient is close to 0, which implies that the problem is very rugged. In this study, we select the degree of PR based on the FDC value to overcome the over-fitting and under-fitting. The FDC value of each sub-problem is computed based on the solutions in $D_g^s$ and a threshold parameter $\varepsilon$ is defined to determine the degree of PR. If $|FDC_g| > \varepsilon$, the second-order PR model is used, otherwise, fifth-order PR model is used.

After the degree of PR model is determined, a specific global PR model is constructed based on the sample points in $D_g^s$ (step 4). This global PR model is able to model the whole solution space and filter out part of the local optimal solutions. And $x_g^*$ can be recognized as the approximant of the real optimal solution to sub-problem $g$ (step 5). We define a small region around $x_g^*$ in step 6. It is believable that the real optimal solution of sub-problem $g$ is within this small region. The new search region is defined according to the following equation:

$$\begin{cases} lb_g^{'} = \dfrac{lb_g}{r} + x_g^* \\ ub_g^{'} = \dfrac{ub_g}{r} + x_g^* \end{cases}, \quad (1)$$

Then, $lb_g^{'}$ and $ub_g^{'}$ are repaired if necessary according to the following equation:

$$\begin{cases} lb_g^{'} = lb_g & \text{if } lb_g^{'} < lb_g \\ ub_g^{'} = ub_g & \text{if } ub_g^{'} > lb_g \end{cases}. \quad (2)$$

**The second layer:** In the second layer, a new $D_g^s$ is firstly generated (steps 7-9). The solutions in $D_g^s$ of the first layer within $[lb_g^{'}, ub_g^{'}]$ are selected and reused (step 7), then $d^s - |O_g|$ new solutions within $[lb_g^{'}, ub_g^{'}]$ are uniformly generated (step 8), and $D_g^s$ is the combination of $O_g$ and $Q_g$ (step 9). After $D_g^s$ is generated, several fifth-order PR models are constructed based on the solutions in $D_g^s$. To construct a fifth-order PR model, the number of training samples must be no less than 6. In order to divide the region into as small as possible sub-regions under the premise that there are at least 6 solutions in each sub-region, we equally divide the search region into $\lfloor d^s / 6 \rfloor$ sub-regions (step 10). Then each $x_g \in D_g^s$ is allocated into the corresponding sub-region (steps 11). If there are less than 6 solutions within a sub-region, the solutions in the neighbor sub-regions can be added to the sub-region to make sure it contains at least 6 training points. After that, a fifth-order PR model is constructed for each sub-region with the corresponding training points in step 13, where $D_g^i$ is the database of the $i$th sub-region. And then, the optimal solution of each PR model within its corresponding sub-region is found out (steps 14), where $lb_i$ and $ub_i$ are lower bound and upper bound of the $i$th sub-region, respectively. Finally, the best one in the $\lfloor d^s / 6 \rfloor$ optimal solutions is selected out (step 15), and if it is better than $x_g^*$, $x_g^*$ is updated (step 16).

## 2.2 Description of RBF-SHADE

After the separable sub-problems are solved, the nonseparable sub-problems will be solved by RBF-SHADE described in Algorithm 3 with the rest of computation resources. It is notable that we still use the real fitness improvement value made by a new sub-solution $x_g$ of a sub-problem to the best overall solution $x^*$ to construct RBF model, and use the RBF model to predict the real fitness improvement value made by a new sub-solution $x_g$ to $x^*$.

Borrowing the experience from [12], in RBF-SHADE, the population $P_g$ records $p$ best sub-solutions obtained so far, and the database $D_g^n$ records $d^n$ most recently real evaluated sub-solutions. It is notable that we record the real fitness improvement values of the individuals in $P_g$ and use these values to update the parameters of SHADE. In Algorithm 3, a RBF model is first generated based on the database $D_g^n$ (step 1). Then, the solutions in $P_g$ and the corresponding trial vectors are all evaluated by the RBF model (steps 2-4). After that, $q$ best trial vectors are selected and reevaluated by the original function in steps 5-6. Then, the parameters of SHADE, $D_g^n$ and $P_g$ are updated (steps 7-12). Finally, if a better solution is found, $x^*$ is updated (steps 14-15), and the fitness improvements of the sub-solutions in $D_g^n$ and $P_g$ are also updated (steps 16-17). For the generation of the trial vectors and concrete update rule of the parameters of SHADE, readers can refer to [9].

| **Algorithm 3**: $(P_g, D_g^n, x^*) \leftarrow RBF - SHADE(P_g, D_g^n, x^*, g)$ |
|---|
| 1. Build a RBF model for the $g$th sub-problem with $D_g^n$; |
| 2. **for** each sub-solution $x_g^i \in P_g$ **do** |
| 3. $\quad$ Generate a trial vector $u_g^i$; |
| 4. $\quad$ Evaluate $x_g^i$ and $u_g^i$ with $\overline{e}(x_g^i)$ and $\overline{e}(u_g^i)$ provided by RBF; |
| 5. Select $q$ best trial vectors from the group of $u_g^i, i = 1, 2, \cdots, p$ and store them into $Q_g$; |
| 6. Reevaluate each trial vector $u_g^i \in Q_g$ with $e(u_g^i)$ and modify $\overline{e}(u_g^i) \leftarrow e(u_g^i)$; |
| 7. Update the parameters of SHADE based on $\overline{e}(x_g^i)$ and $\overline{e}(u_g^i)$; |
| 8. Replace the $q$ oldest sub-solutions in $D_g^n$ with the trial vectors in $Q_g$; |
| 9. **for** each $u_g^i \in Q_g$ **do** |
| 10. $\quad$ Find out the worst sub-solution $x_g^w$ in $P_g$; |
| 11. $\quad$ **if** $e(x_g^w) < e(u_g^i)$ **then** |
| 12. $\quad\quad$ Delete $x_g^w$ from $P_g$ and insert $u_g^i$ into $P_g$; |
| 13. Find out the best sub-solution $x_g^b$ in $P_g$; |
| 14. **if** $e(x_g^b) > 0$ **then** |
| 15. $\quad$ Update $f(x^*) \leftarrow f(x^* | x_g^b)$, $x^* \leftarrow x^* | x_g^b$; |
| 16. $\quad$ **for** each sub-solution $x_g^i \in D_g^n \cup P_g$ **do** |
| 17. $\quad\quad$ Update $e(x_g^i) \leftarrow e(x_g^i) - e(x_g^b)$; |
| 18. Return $P_g, D_g^n, x^*$; |

The aim of steps 16-17 is to make the solutions in $D_g^n$ and $P_g$ commensurable. It is clearly that the sub-solutions in archive $D_g^n$ and population $P_g$ are generally introduced in different iterations, during which the context vectors, i.e., $x^*$, may change. Without fine intervention, this would make the fitness improvements of these sub-solutions incommensurable. A straightforward way to tackle this issue is to reevaluate current sub-solutions in $D_g^n$ and $P_g$ if the context vector is really updated, but this will consume extra computation resources. To avoid this, once a better context vector $x^*|x_g^b$ is found, RBF-SHADE updates the fitness improvements of all the sub-solutions in current $D_g^n$ and $P_g$ according to the way in step 17.

From Algorithm 3, it can be seen that only $q/p \times 100\%$ of sub-solutions in RBF-SHADE need to be evaluated by the original simulation model, which significantly reduces the number of real evaluations since $q$ is generally much less than $p$. And it can also be seen from Algorithm 2 that for a 1-dimensional sub-problem $g$, only $2d^s$ FEs are required. Usually, $d^s$ is a small integer so that the 1-dimensional sub-problems only require a little FEs. As explained above, ASMCC has the potential to obtain better solutions with much fewer FEs.

## 3 EXPERIMENTAL STUDIES

### 3.1 Experimental Settings

The CEC 2010 benchmark suite which contains 20 LSOPs was employed in our experiments [13]. All these benchmark functions are minimization problems of 1000 dimensions. It is known that there are no separable variables in $F_{14}$-$F_{20}$, therefore they were excluded from our experiments. Two decomposition methods were used in our experiments. One is the ideal decomposition which groups variables according to the prior knowledge of a benchmark function. The other is the recently developed vector-growth decomposition algorithm method named VGDA-D [14]. VGDA-D is a vector based decomposition method which can decompose LSOPs with high accuracy using just a little number of FEs.

As suggested by [13], most existing CC algorithms take a maximum number of $3.0 \times 10^6$ FEs as the termination condition of a run. To show the superiority of ASMCC, it only employed 10 percent of the suggested computation resource, i.e., a maximum number of $3.0 \times 10^5$ FEs, as the default termination condition of a run. And the result of each algorithm on a function was calculated based on 25 independent runs.

There are several parameters in ASMCC. For the two-layer PR search process, there are 3 parameters, including the size $d^s$ of $D_g^s$, the threshold parameter $\varepsilon$, and the parameter $r$ to determine the new region. Based on our prior experiments, we suggest setting $d^s$ to 100, $\varepsilon$ to 0.8. As for $r$, our pilot experiments shown that, if $r$ is set to 10 and 15 when the fifth-order PR model and second-order PR model are used, respectively, the new search region can cover almost all of the optimal sub-solutions of different LSOPs. So we suggest setting $r$ to 10 and 15 when fifth-order and second-order PR model is used, respectively. As for RBF-SHADE, three are also 3 parameters, including the size $d^n$ of $D_g^n$, the population size $p$ of SHADE and the number of reevaluated solutions $q$. For the achieve size $d^n$, it is set to $5D$ according to the suggestion given in [15], where $D$ is the dimensionality of the corresponding sub-problem $g$. For the population size $p$, it has been investigated much in the original SHADE [9], and it is revealed that the algorithm performs well on most small and medium scale problems when the population size is set to 100. This conclusion was also verified by our pilot experiments. Accordingly, the population size was set to 100 for RBF-SHADE. For the number of reevaluated solutions $q$, our prior experiments suggest setting it to 10.

### 3.2 Comparison between ASMCC and other CC Algorithms under the Ideal Decomposition

To show the effectiveness of ASMCC, we compared it with three other CC algorithms under the ideal decomposition in this sub-section. At first, in order to verify the efficiency of the RBF model, we implemented a CC algorithm which employs the two-layer PR search process and SHADE to solve the separable sub-problems and nonseparable sub-problems, respectively, and name it PS-CC. Then, in order to verify the efficiency of the two-layer PR search process, we implemented a traditional CC algorithm which employs SHADE as optimizer and name it SHADE-CC. When adopting ideal decomposition, the three algorithms are represented as SHADE-CC-I, PS-CC-I and ASMCC-I. Moreover, we also compared the three algorithms with an existing CC algorithm developed in [16] with a name of CC-I. Different from SHADE-CC, CC-I uses another efficient DE variant named SaNSDE as optimizer.

To ensure the fairness of the comparison, the parameters in SHADE-CC-I and PS-CC-I were set to the same values as the corresponding ones in ASMCC-I. Table 1 summarizes the results obtained by SHADE-CC-I, PS-CC-I and ASMCC-I with $3.0 \times 10^5$ FEs and the results obtained by CC-I with $3.0 \times 10^6$ FEs. It is necessary to mention that the results of CC-I is directly taken from [16]. And to statistically analyze the performance of the three competitors, we employed Cohen's $d$ effect size [17] to quantify the difference among the average fitness values (FVs) obtained by them. If a result in Table 1 is judged to be better than, worse than, or similar to the corresponding one obtained by ASMCC-I, it is marked with "+", "−", and "≈", respectively.

Following results can be concluded from Table 1:

1. SHADE-CC-I outperforms CC-I on 12 out of total 13 functions. Based on this result, it can be concluded that SHADE has an edge over SaNSDE for LSOPs under the CC framework since the difference between SHADE-CC-I and CC-I mainly lies in the optimizer and the former only consumes 10% of real FEs consumed by the latter.

2. In order to verify the efficiency of the two-layer PR search process, SHADE-CC-I and PS-CC-I are compared. It can be seen that PS-CC-I outperforms SHADE-CC-I on all of the 18 benchmark functions when the same number of FEs are used. The superiority of PS-CC-I over SHADE-CC-I reveals that the two-layer PR search process is really feasible and efficient.

3. In order to verify the efficiency of the RBF model, AMSCC-I and PS-CC-I are compared. It can be observed that AMSCC-I achieves better results than PS-CC-I. It outperforms PS-CC-I on 8 functions and obtains similar results on 3 functions out of total 13 functions. This result indicates that the RBF model makes sense and can help the algorithm obtain better solution. The two algorithms obtain similar results on functions $F_1$-$F_3$ because there are no nonseparable variables in these functions and the two algorithms solve them by the same way. ASMCC-I performs worse than PS-CC-I on functions $F_6$ and $F_{11}$. The main reason consists in that for $F_6$ and $F_{11}$, they take Ackley function as the basic function, whose fitness landscape is nearly a plateau in the solution region close to the global optimum and the optimum is located in a very narrow region near the origin [13]. Then it is very difficult to build accurate enough RBF models for the nonseparable sub-problems in $F_6$ and $F_{11}$ with a limited number of samples, which restricts the performance of ASMCC-I on them so that PS-CC-I can find better solutions.

**Table 1: The average FVs ± standard deviations obtained by CC-I, SHADE-CC-I, PS-CC-I and ASMCC-I on CEC 2010 functions**

| F | CC-I | SHADE-CC-I | PS-CC-I | ASMCC-I |
|---|---|---|---|---|
| $F_1$ | 3.50e+11 ± 2.0e+10$^-$ | 1.05e+06 ± 1.03e+05$^-$ | **7.05e-14 ± 1.59e-15**$^*$ | 7.05e-14 ± 1.59e-15 |
| $F_2$ | 9.40e+03 ± 2.1e+02$^-$ | 6.51e+03 ± 5.69e+01$^-$ | **7.32e-06 ± 4.44e-07**$^*$ | 7.32e-06 ± 4.44e-07 |
| $F_3$ | 2.00e+01 ± 4.4e−02$^-$ | 1.52e+01 ± 1.76e-01$^-$ | **5.61e-03 ± 2.51e-02**$^*$ | 5.61e-03 ± 2.51e-02 |
| $F_4$ | 3.40e+14 ± 7.5e+13$^-$ | 6.92e+13 ± 1.15e+13$^-$ | 6.05e+11 ± 2.66e+11$^-$ | **8.67e+10 ± 3.94e+10** |
| $F_5$ | 4.90e+08 ± 2.4e+07$^-$ | 4.01e+08 ± 1.91e+07$^-$ | 1.26e+08 ± 1.46e+07$^-$ | **1.12e+08 ± 2.61e+07** |
| $F_6$ | 1.10e+07 ± 7.5e+05$^-$ | 1.05e+06 ± 1.81e+05$^-$ | **1.11e-02 ± 3.54e-02**$^+$ | 3.87e+05 ± 7.55e+05 |
| $F_7$ | 7.70e+10 ± 9.6e+09$^-$ | 2.68e+10 ± 5.27e+09$^-$ | 7.91e+04 ± 8.93e+04$^-$ | **1.63e-03 ± 2.62e-03** |
| $F_8$ | 1.80e+14 ± 9.3e+13$^-$ | 3.46e+09 ± 2.39e+09$^-$ | 3.51e+07 ± 2.51e+07$^-$ | **9.57e+05 ± 1.74e+06** |
| $F_9$ | 9.40e+08 ± 7.1e+07$^-$ | 6.13e+08 ± 4.14e+07$^-$ | 3.34e+08 ± 2.29e+07$^-$ | **1.14e+07 ± 1.55e+06** |
| $F_{10}$ | 4.80e+03 ± 6.7e+01$^-$ | 7.29e+03 ± 8.80e+01$^-$ | 3.75e+03 ± 5.53e+01$^-$ | **1.10e+03 ± 7.03e+01** |
| $F_{11}$ | 4.10e+01 ± 1.5e+00$^-$ | 2.74e+01 ± 8.35e-01$^-$ | **9.00e-01 ± 1.16e-01**$^+$ | 6.00e+00 ± 2.33e+00 |
| $F_{12}$ | 4.90e+05 ± 3.4e+04$^-$ | 2.98e+05 ± 1.20e+04$^-$ | 1.76e+05 ± 8.98e+03$^-$ | **1.82e+03 ± 1.27e+03** |
| $F_{13}$ | 1.50e+07 ± 4.1e+06$^-$ | 3.28e+04 ± 5.77e+03$^-$ | 3.76e+03 ± 1.46e+03$^-$ | **6.47e+02 ± 1.97e+02** |
| +/≈/− | 0/0/13 | 0/0/13 | 2/3/8 | − |
| Ranking | 3.9231 | 3.0769 | 1.7692 | 1.2308 |

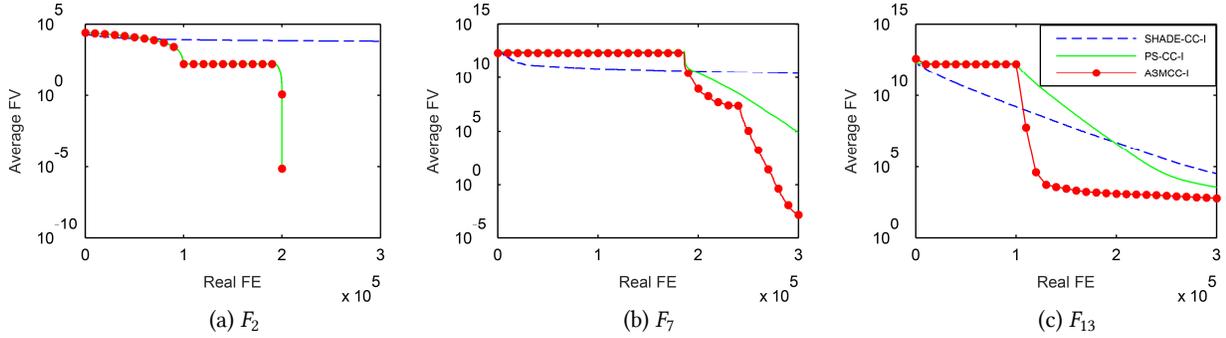

**Figure 1: The evolution trends of average FVs obtained by SHADE-CC-I, PS-CC-I and ASMCC-I on $F_2$, $F_7$ and $F_{13}$.**

4. The last row of Table 1 lists the ranking of the four algorithms according to Friedman test, from which it can be concluded that ASMCC-I performs best, followed by PS-CC-I and SHADE-CC-I, whereas the traditional CC-I is definitely defeated by the other three algorithms. This result demonstrates that the ASMCC framework is rather successful.

In order to examine the evolution characteristics of ASMCC, Fig. 1 compares the evolution curves of the average FVs obtained by SHADE-CC-I, PS-CC-I and ASMCC-I, where functions $F_2$, $F_7$ and $F_{13}$ are taken as examples. From Fig. 1, it can be seen that both ASMCC-I and PS-CC-I obtain the optimum of the fully separable function $F_2$ with only $2.0 \times 10^5$ FEs. With respect to partially separable functions $F_7$ and $F_{13}$, two phenomena can be observed. On one hand, PS-CC-I and AMSCC-I perform slightly worse than SHADE-CC-I at the early stage of the evolution process on $F_7$ and $F_{13}$. The reason mainly consists in that there is an imbalance among the contributions of various sub-problems to the global fitness of the individuals. In the two functions, the nonseparable sub-problems contribute more to the global fitness than the separable sub-problems. So that in the early stage, when the two algorithms focus on finding the optimal solutions of the separable sub-problems, although they find very good solutions, the best FV didn't change much because it is mainly determined by the solutions of the nonseparable sub-problems. When the algorithms begin to optimize the nonseparable sub-problems, better FVs will be achieved. On the other hand, ASMCC-I and PS-CC-I yield better solutions than SHADE-CC-I after the separable sub-problems are optimized and keep a better evolution trend until all the available computation resources are exhausted.

Table 2: The average FVs ± standard deviations obtained by MA-SW-Chains, SHADE-CC-D, PS-CC-D and ASMCC-D on CEC2010 functions

| $F$ | MA-SW-Chains | SHADE-CC-D | PS-CC-D | ASMCC-D |
|---|---|---|---|---|
| $F_1$ | **2.10e−14 ± 1.99e−14**$^+$ | 1.16e+06 ± 8.08e+04$^-$ | 7.05e-14 ± 1.56e-15$^*$ | 7.05e-14 ± 1.53e-15 |
| $F_2$ | 8.10e+02 ± 5.88e+01$^-$ | 6.51e+03 ± 7.29e+01$^-$ | **7.19e-06 ± 5.10e-07**$^*$ | 7.19e-06 ± 5.11e-07 |
| $F_3$ | **7.28e−13 ± 3.40e−13**$^+$ | 1.49e+01 ± 3.69e-01$^+$ | 1.50e+01 ± 3.69e-01$^+$ | – |
| $F_4$ | 3.53e+11 ± 3.12e+10$^-$ | 7.23e+13 ± 1.34e+13$^-$ | 8.21e+11 ± 3.30e+11$^-$ | **7.86e+10 ± 2.90e+10** |
| $F_5$ | 1.68e+08 ± 1.04e+08$^-$ | 4.04e+08 ± 1.83e+07$^-$ | 1.27e+08 ± 1.49e+07$^-$ | **1.18e+08 ± 1.87e+07** |
| $F_6$ | 8.14e+04 ± 2.84e+05$^+$ | 1.08e+06 ± 2.43e+05$^-$ | **2.36e-02 ± 5.36e-02**$^+$ | 9.18e+05 ± 1.18e+06 |
| $F_7$ | 1.03e+02 ± 8.70e+01$^-$ | 2.51e+10 ± 4.73e+09$^-$ | 1.78e+05 ± 3.72e+05$^-$ | **2.26e-03 ± 7.02e-03** |
| $F_8$ | 1.41e+07 ± 3.68e+07$^-$ | 3.50e+09 ± 1.79e+09$^-$ | 3.08e+07 ± 1.72e+07$^-$ | **7.17e+05 ± 1.48e+06** |
| $F_9$ | 1.41e+07 ± 1.15e+06$^-$ | 6.52e+08 ± 4.83e+07$^-$ | 3.65e+08 ± 2.56e+07$^-$ | **1.17e+07 ± 1.07e+06** |
| $F_{10}$ | 2.07e+03 ± 1.44e+02$^-$ | 7.35e+03 ± 5.69e+01$^-$ | 3.80e+03 ± 4.72e+01$^-$ | **1.11e+03 ± 9.68e+02** |
| $F_{11}$ | 3.80e+01 ± 7.35e+00$^-$ | **1.63e+01 ± 3.41e-01**$^+$ | **1.63e+01 ± 4.21e-01**$^+$ | 2.33e+01 ± 4.44e+00 |
| $F_{12}$ | **3.62e−06 ± 5.92e−07**$^+$ | 3.09e+05 ± 1.71e+04$^-$ | 2.00e+05 ± 1.13e+04$^-$ | 1.70e+03 ± 7.31e+02 |
| $F_{13}$ | 1.25e+03 ± 5.72e+02$^-$ | 1.25e+05 ± 2.11e+04$^-$ | 1.03e+04 ± 2.20e+03$^-$ | **8.01e+02 ± 3.50e+02** |
| +/≈/− | 4/0/9 | 2/0/11 | 3/2/8 | – |
| Ranking | 2.1667 | 3.7500 | 2.5000 | 1.5833 |

## 3.3 Comparison among ASMCC, other CC Algorithms under VGDA-D and a Non-CC Algorithm

To evaluate the performance of ASMCC more comprehensively, we further tested it coupled with VGDA-D which is an efficient decomposition method developed recently. For the convenience of description, we abbreviate ASMCC, PS-CC and SHADE-CC which adopt VGDA-D to ASMCC-D, PS-CC-D and SHADE-CC-D. Moreover, we also compared the three CC algorithms with a non-CC algorithm, i.e., MA-SW-Chains [18], which is a memetic algorithm and was ranked the first in the IEEE CEC 2010 competitions on LSOP. Table 2 summarizes the results obtained by SHADE-CC-D, PS-CC-D and ASMCC-D with $3.0 \times 10^5$ FEs, and the results obtained by MA-SW-Chains with $3.0 \times 10^6$ FEs. It is necessary to mention that the FEs consumed during the decomposition process were counted into the allowed maximum FE number, and the results of MA-SW-Chains are directly taken from [18].

For the CC algorithms SHADE-CC, PS-CC and ASMCC, Table 2 shows the same trend as Table 1 except for $F_3$ and $F_{11}$. It is because that VGDA-D didn't get the right decomposition on functions $F_3$ and $F_{11}$, it decomposed $F_3$ into a 1000-dimensional nonseparable sub-problem and $F_{11}$ into ten 50-dimensional and one 500 dimensional nonseparable sub-problems [14]. For PS-CC-D, there are no separable sub-problems in $F_3$ and $F_{11}$ so that the two-layer PR search process makes no sense. It means that PS-CC-D degenerated into SHADE-CC-D when solving $F_3$ and $F_{11}$ so they get similar results. For ASMCC-D, it is impossible to construct accuracy enough RBF models for the 1000-dimensional sub-problem in $F_3$ so $F_3$ is excluded from ASMCC-D. As for $F_{11}$, the 500-dimensional sub-problem greatly restricts the performance of the RBF-SHADE algorithm in ASMCC-D so that it is defeated by PS-CC-D.

As for MA-SW-Chains, it outperforms SHADE-CC-D and PS-CC-D on 12 and 9 functions, respectively. But it is defeated by ASMCC-D on 9 functions. This result further shows that the two-layer surrogate search process for the separable sub-problems and the SAEA for the nonseparable sub-problems are really feasible and efficient.

The last row of Table 2 also lists the ranking of the four algorithms according to Friedman test when $F_3$ is excluded. From which it can be concluded that ASMCC-D performs best, followed by MA-SW-Chains and PS-CC-D, whereas the traditional SHADE-CC-D is definitely defeated by the other three algorithms. This result proves that the proposed ASMCC framework is suitable for practical decomposition strategy, and the proposed ASMCC algorithm is highly competitive in solving LSOPs.

## 4 CONCLUSIONS

In this paper, a novel ASMCC framework is proposed to solve LSOP. ASMCC is characterized by solving different sub-problems of CC in different ways. It solves the separable sub-problems by further divided them into a number of 1-dimensional sub-problems and then solves these 1-dimensional sub-problems by a two-layer surrogate search process, while solving the nonseparable sub-problems by a specific SAEA. By this way, ASMCC greatly reduces the requirement on the number of real FEs and improves the search efficiency of CC. Experimental results on CEC 2010 benchmark suite demonstrate that ASMCC is compatible with different decomposition methods, and has an edge over the traditional CC and non-CC algorithms.

The ASMCC algorithm presented in this paper is only adopted to solve the LSOPs which have separable variables. Our future work will focus on developing ASMCC algorithms which can solve all kinds of LSOPs. Moreover, we will further verify the efficiency of ASMCC on other benchmark functions and some real world problems.

## A APPENDICES

In this section, we present the detailed description of FDC and the RBF model used in this paper.

## A.1 Fitness Distance Correlation

The modified FDC described in [11] is used in this paper. It is described as follows:

$$FDC = \frac{\sum_{i=1}^{n}(f_i - \bar{f})(d_i^* - \bar{d}^*)}{\sqrt{\sum_{i=1}^{n}(f_i - \bar{f})^2}\sqrt{\sum_{i=1}^{n}(d_i^* - \bar{d}^*)^2}}, \quad (3)$$

where $f_i \in F = \{f_1, f_2, ... f_n\}$ is the fitness values of the $n$ sample points, $\bar{f}$ is the mean of $F$, $d_i^* \in D^* = \{d_1^*, d_2^*, ..., d_n^*\}$ is the distance between the $i$th solution and the best solution in the current sample set, $\bar{d}^*$ is the mean of $D^*$.

## A.2 Radial Basis Function

This paper uses the RBF model which is used in [19]. Given $d$ training samples $t^1, t^2, \cdots, t^d \in \mathbb{R}^D$ for a sub-problem $g$ of $D$ dimensions, the evaluation value provided by RBF for a new sub-solution $x_g \in \mathbb{R}^D$ can be represented as

$$\bar{e}(x_g) = \sum_{i=1}^{d} \omega_i \phi(\|x_g - t^i\|) + \beta^T x_g + \alpha, \quad (4)$$

where $\|x_g - t^i\|$ denotes the distance between the two solutions and the Euclidean distance is generally employed, $\phi(\cdot)$ denotes the basis function, this study adopts cubic basis function, i.e., $\phi(r) = r^3$, $\omega = (\omega_1, \omega_2, \cdots, \omega_d)^T \in \mathbb{R}^d, \beta \in \mathbb{R}^D$, and $\alpha \in \mathbb{R}$ are corresponding parameters. As for $\beta^T x_g + \alpha$, it is a polynomial tail appended to the standard RBF.

## ACKNOWLEDGMENTS

This work was supported in part by the National Natural Science Foundation of China under Grant 61105126 and in part by the Postdoctoral Science Foundation of China under Grants 2014M560784 and 2016T90922.